\documentclass{article}

\usepackage{microtype}
\usepackage{graphicx}
\usepackage{subfigure}
\usepackage{booktabs} 
\usepackage{multirow}
\usepackage{ulem}
\usepackage{tabularx}
\usepackage{tabu}

\usepackage{hyperref}
\usepackage{amsmath,amsfonts,bm}
\usepackage{color}

\usepackage[accepted]{icml2021}

\icmltitlerunning{Towards Defending against Adversarial Examples via Attack-Invariant Features}

\begin{document}

\twocolumn[
\icmltitle{Towards Defending against Adversarial Examples \\ via Attack-Invariant Features}



\icmlsetsymbol{correspond}{$\dagger$}

\begin{icmlauthorlist}
\icmlauthor{Dawei Zhou}{xd,tml}
\icmlauthor{Tongliang Liu}{tml}
\icmlauthor{Bo Han}{hkbu}
\icmlauthor{Nannan Wang}{xd,correspond}
\icmlauthor{Chunlei Peng}{wx}
\icmlauthor{Xinbo Gao}{cu}

\end{icmlauthorlist}

\icmlaffiliation{xd}{State Key Laboratory of Integrated Services Networks, School of Telecommunications Engineering, Xidian University}
\icmlaffiliation{hkbu}{Department of Computer Science, Hong Kong Baptist University}
\icmlaffiliation{wx}{State Key Laboratory of Integrated Services Networks, School of Cyber Engineering, Xidian University}
\icmlaffiliation{cu}{Chongqing Key Laboratory of Image Cognition, Chongqing University of Posts and Telecommunications}
\icmlaffiliation{tml}{Trustworthy Machine Learning Lab, School of Computer Science, The University of Sydney}

\icmlcorrespondingauthor{Nannan Wang}{nnwang@xidian.edu.cn}

\icmlkeywords{Machine Learning, ICML}

\vskip 0.3in
]



\printAffiliationsAndNotice{}  

\begin{abstract}
Deep neural networks (DNNs) are vulnerable to adversarial noise. Their adversarial robustness can be improved by exploiting adversarial examples. However, given the continuously evolving attacks, models trained on seen types of adversarial examples generally cannot generalize well to unseen types of adversarial examples. To solve this problem, in this paper, we propose to remove adversarial noise by learning generalizable invariant features across attacks which maintain semantic classification information. Specifically, we introduce an adversarial feature learning mechanism to disentangle invariant features from adversarial noise. A normalization term has been proposed in the encoded space of the attack-invariant features to address the bias issue between the seen and unseen types of attacks. Empirical evaluations demonstrate that our method could provide better protection in comparison to previous state-of-the-art approaches, especially against \textit{unseen} types of attacks and \textit{adaptive} attacks.

\end{abstract}

\section{Introduction}
Deep neural networks (DNNs) have been widely utilized in many fields, such as image processing \cite{lecun1998gradient,he2016deep,Zagoruyko2016WRN,simonyan2014very,2017Mask} and natural language processing \citep{sutskever2014sequence}. 
However, DNNs are found to be vulnerable to adversarial examples which are crafted by adding imperceptible but adversarial noise on natural examples \cite{goodfellow2014explaining,szegedy2013intriguing,shen2017ape,liao2018defense,ma2018characterizing}. The vulnerability of DNNs poses serious risks in many security-sensitive applications such as face recognition \cite{xu2020adversarial} and autonomous driving \cite{eykholt2018robust}.

Existing methods show that the adversarial robustness of target models can be enhanced by exploiting adversarial examples, e.g., employing the adversarial examples as additional training data \cite{goodfellow2014explaining,tramer2017ensemble, wu2019defending}. However, focusing on the seen types of adversarial examples in the finite training data would cause the defense method to overfit the given types of adversarial noise and lack generalization or effectiveness against unseen types of attacks. Note that there are widespread or even unprecedented types of attacks in the real world. This motivates us to design a defense method that could handle different and unseen types of adversarial examples.

\begin{figure}[t]
\vskip 0.2in
\begin{center}
\centerline{\includegraphics[width=\columnwidth]{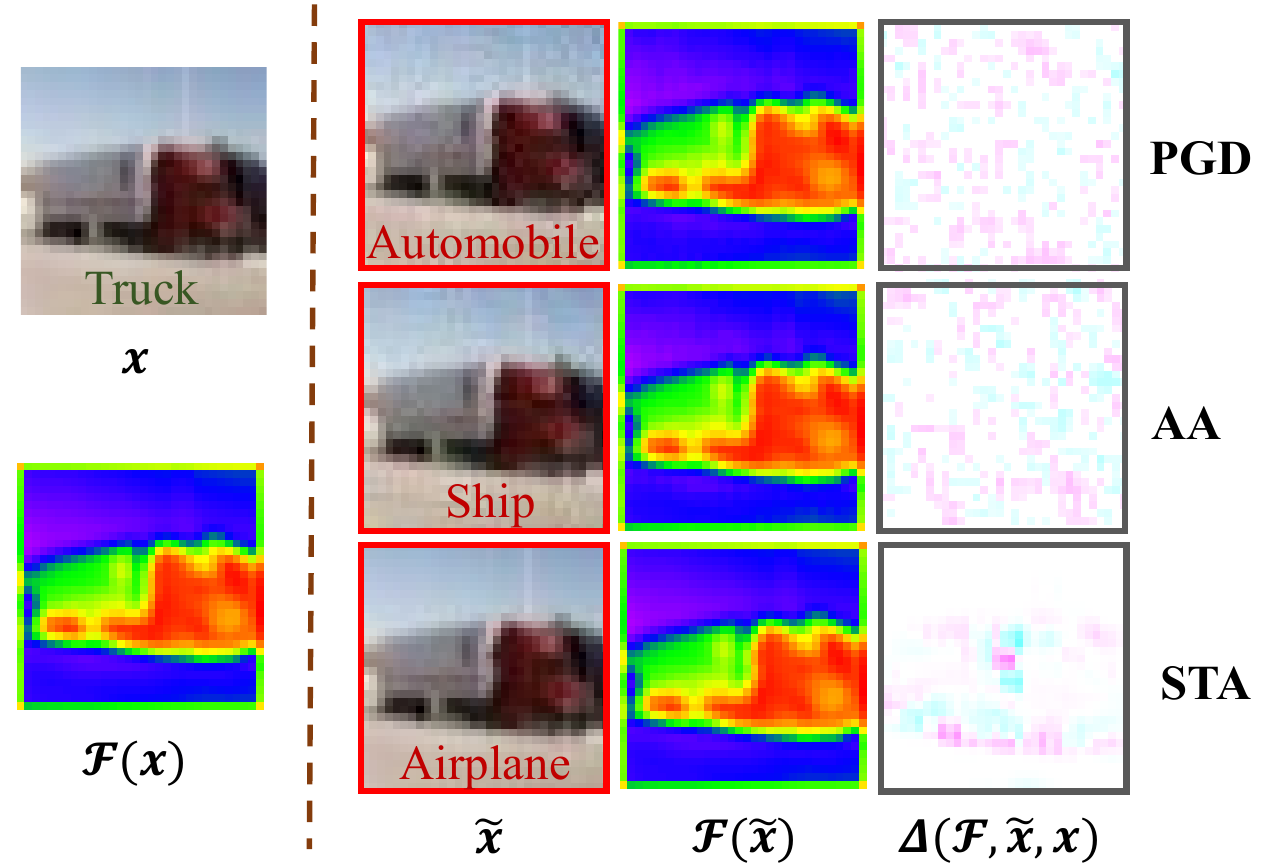}}
\caption{A visual illustration of the natural example ($x$), adversarial example ($\tilde{x}$), latent feature ($\mathcal{F}(\cdot)$) and attack-specific feature ($\Delta(\mathcal{F},\tilde{x},x)=[\mathcal{F}(\tilde{x})\text{ - }\mathcal{F}(x)]\times10^{4}$). The latent feature is extracted from the first ReLu layer of the ResNet-110 model \citep{he2016deep}. Different types of attacks (i.e., PGD \cite{madry2017towards}, AA \cite{croce2020reliable} and STA \cite{xiao2018spatially}) generally only modify tiny information and their adversarial examples sufficiently retain invariant features from the natural examples.}
\label{fig2}
\end{center}
\vskip -0.22in
\end{figure}

Cognitive science gives us an inspiration to solve this problem. Specifically, it shows that we are able to identify human faces even if the faces show different or even unseen expressions, because our brain is good at extracting invariant facial features \cite{mishkin1982contribution,kanwisher1997fusiform}. Similarly, we human cannot easily distinguish natural examples and adversarial examples, because we focus on the invariant features which represent semantic classification information and ignore the adversarial noise. Note that adversarial examples are designed to retain the invariant features so that we human could not identify the adversarial examples in advance, e.g., by constraining the adversarial noise to be small or non-suspicious \cite{goodfellow2014explaining,gilmer2018motivating}. We name such invariant features as \textit{attack-invariant features} (AIF).

In this paper, we propose an \textit{adversarial noise removing network} (ARN) to restore the natural data by exploiting AIF, which is able to defend against unseen types of adversarial examples. In a high level, we design an autoencoder-based framework, which divides the adversarial noise removing into learning AIF and restoring natural examples from AIF. Specifically, we introduce a pair of encoder and discriminator in an adversarial feature learning manner for disentangling AIF from adversarial noise. The discriminator is devoted to distinguish attack-specific information (e.g., attack type label) from the encoded AIF space, while the encoder aims to learn features which are indistinguishable for the discriminator. By iterative optimization, the attack-specific information will be removed and the invariant features across attacks will be retained. 

Note that the adversarial examples used in the training procedure are often biased because the widespread types of attacks in the real world are very diverse. For example, as shown in Figure~\ref{fig2}, the autoattack (AA) method based adversarial example \cite{croce2020reliable} and the projected gradient descent (PGD) method based adversarial example \cite{madry2017towards} are similar; while they are quite different from the spatial transform attack (STA) based adversarial example \cite{xiao2018spatially}. If we do not handle the bias problem, the learned AIF may work well for the attacks or similar types of attacks used in the training procedure, but may have poor generalization ability for some unseen types of attacks whose perturbations are significantly diverse from those in seen types of attacks \cite{li2018domain,makhzani2015adversarial}. To address the bias issue, we impose a normalization term in the encoded space of AIF to match the feature distribution of each type of attack to a multivariate Gaussian prior distribution \cite{makhzani2015adversarial,kingma2013auto}. By this design, the learned AIF is expected to generalize well to widespread unseen types of attacks.

To restore the original natural examples from AIF, a decoder is trained by minimizing the gap between the synthesized examples and the natural examples in the pixel space. Achieved by jointly optimizing the encoder and decoder for learning AIF, our ARN could provide more superior protection against unseen types of attacks compared to previous methods. This will be empirically verified on pixel-constrained and spatially-constrained attacks in Section~\ref{section4.2}. Furthermore, additional evaluations on cross-model defenses and adversarial example detection in Section~\ref{section4.3} further show the effectiveness of ARN. The main contributions in this paper are as follows:
\begin{itemize}
    \item Adversarial examples typically have shared invariant features even if they are crafted by unseen types of attacks. We propose an \textit{adversarial noise removing network} (ARN) to effectively remove adversarial noise by exploiting \textit{attack-invariant features} (AIF). 
    \item To handle the bias issue of the adversarial examples available in the training procedure, we design a normalization term in the encoded AIF space to enhance its generalization ability to unseen types of attacks.
    \item Empirical experiments show that our method presents superior effectiveness against both pixel-constrained and spatially-constrained attacks. Particularly, the success rates of unseen types of attacks and adaptive attacks are reduced in comparison to previous state-of-the-art approaches.
\end{itemize}

The rest of this paper is organized as follows. In Section~\ref{section2}, we briefly review related work on attacks and defenses. In Section~\ref{section3}, we describe our defense method and present its implementation. Experimental results against both pixel-constrained and spatially-constrained attacks are provided in Section~\ref{section4}. Finally, we conclude this paper in Section~\ref{section5}. 

\section{Related work}
\label{section2}

\begin{figure*}[t]
\vskip 0.2in
\begin{center}
\centerline{\includegraphics[width= \textwidth]{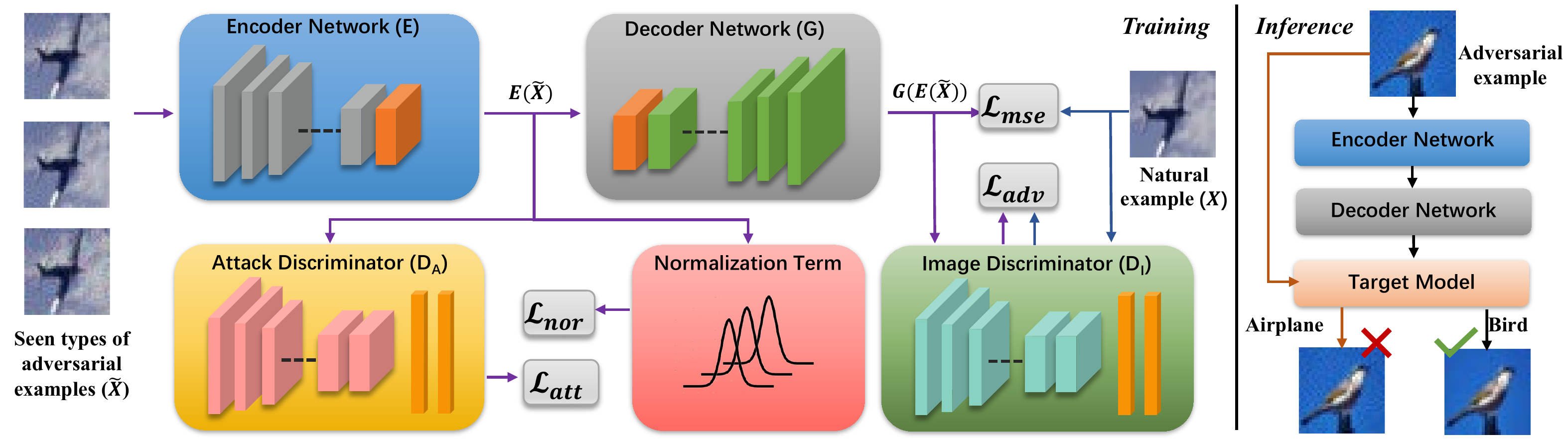}}
\caption{A visual illustration of our \textit{adversarial noise removing network} (ARN). Our main idea is to restore natural examples by exploiting invariant features. ARN is composed of an encoder network and a decoder network. The encoder network learns attack-invariant features (AIF) via an adversarial feature learning mechanism and a normalization term. The decoder is trained to restore natural examples from AIF via a pixel similarity metric and an image discriminator.}
\label{fig3}
\end{center}
\vskip -0.2in
\end{figure*}

\textbf{Attacks:} The seminal work of Szegedy \textit{et al.} \yrcite{szegedy2013intriguing} first proposed adversarial examples that can mislead DNNs. Adversarial examples can be crafted by adding adversarial noise following the direction of adversarial gradients. Attacks based on this strategy include fast gradient sign method (FGSM) \cite{goodfellow2014explaining}, the strongest first-order information based projected gradient descent (PGD) method \cite{madry2017towards}, the Jacobian-based saliency map attack (JSMA) method \cite{papernot2016limitations}. The autoattack (AA) method \citep{croce2020reliable} forms a parameter-free, computationally affordable and user-independent ensemble of attacks. The adversarial noise crafted by these attacks is typically bounded by a small norm-ball $\|\cdot\|_{p} \leq \epsilon$, so that their adversarial examples can be perceptually similar to natural examples. In addition, optimization-based attacks, such as Carlini and Wagner (CW) method \cite{carlini2017towards} and decoupling direction and norm (DDN) method \cite{rony2019decoupling}, minimize the adversarial noise as part of the optimization objectives. The above attacks directly modify the pixel values on the whole sample without considering semantics of objectives, e.g., shape and posture. They are named as \textit{pixel-constrained attacks}. In addition, there are also \textit{spatially-constrained attacks} which focus on mimicking non-suspicious vandalism via geometry and spatial transformation or physical modifications. These attacks include faster wasserstein attack (FWA) \cite{wu2020stronger}, spatial transform attack (STA) \cite{xiao2018spatially} and robust physical perturbations (RP2) \cite{eykholt2018robust}.

\textbf{Defenses:} Adversarial training (AT) is a widely used strategy for defending against adversarial noise by augmenting the training data with adversarial examples, such as PGD based adversarial training method ($\text{AT}_{PGD}$) \cite{madry2017towards} and defending against occlusion attacks (DOA) \citep{wu2019defending} method. In addition, input processing based methods have also been proposed to defend against attacks. They aim to pre-process input data for mitigating the aggressiveness of adversarial noise. For example, Jin \textit{et al.} \yrcite{shen2017ape} proposed APE-G to back adversarial examples close to natural examples via a generative adversarial network. Liao \textit{et al.} \yrcite{liao2018defense} utilized a high-level representation guided denoiser (HGD) as a pre-processing step to remove adversarial noise. HGD used the class labels predicted by a target model to supervise the training of an end-to-end denoiser. Compared with the above defenses, we design an input processing based model that remove adversarial noise by learning attack-invariant features, instead of directly relying on learning a function which maps seen types of adversarial examples to the perceptual space of natural examples. In addition, the method in \cite{xu2017feature} shows that reducing the color bit depth of an adversarial example could reduce its attack success rate, but the method may make processed examples lose some useful natural features. Our method brings adversarial examples close to the natural examples without causing human-observable loss. The defense in \cite{xie2019feature} focuses on denoising the perturbations in the feature maps on internal layers of the target model by modifying the target models’ architectures. Differently, our method aims to disentangle natural features from adversarial noise, and use the natural features to generate clean examples. Our defense is a pre-processing based defense, which dose not require the knowledge of the target models and could provide cross-model protection. 

\section{Adversarial noise removing network}
\label{section3}

\subsection{Preliminaries}
In this paper, we aim to design a defense which could provide robust protection against widespread unseen types of attacks. The basic intuition behind our defense is to effectively exploit the invariant features. To this end, we propose the \textit{adversarial noise removing network} (ARN) which eliminates adversarial noise by learning \textit{attack-invariant features} (AIF).  As shown in Figure~\ref{fig3}, our ARN divides the remove of adversarial noise into two steps. The first step is to learn AIF from input examples via an encoder network $E$. The second one is to restore natural examples from AIF via a decoder network $G$. The encoder network $E$ is trained 
by exploiting an attack discriminator $D_A$ and a normalization term, while the decoder network $G$ is trained to minimize the gap between synthetic examples and natural examples via utilizing a pixel similarity metric and an image discriminator $D_I$.

Our defense model can be expressed as $G(E(\tilde{X})$, where $E(\cdot)$ represents the process of learning AIF from input adversarial examples $\tilde{X}$. We use $\tilde{X}_{k}=\left[\tilde{x}_{k_{1}},\tilde{x}_{k_{2}}, \ldots, \tilde{x}_{k_N}\right]^{\top}$ to denote adversarial data for the $k$-$th$ type of attack, where $k_N$ is the number of adversarial examples crafted by the $k$-$th$ type of attack. The natural data are denoted by ${X}=\left[{x}_1,{x}_2, \ldots, {x}_N\right]^{\top}$, where $N$ is the number of the natural examples. 

\subsection{Learning Attack-invariant Features}
We propose a hybrid objective function to train our ARN to learn AIF. The objective function consists of two terms that we explain below:


\textbf{Adversarial feature learning:} To remove attack-specific information by disentangling invariant features from adversarial noise, we distinguish the different types of adversarial noise in the resulting encoded feature space. More precisely, we introduce an attack discriminator $D_A$ to form an adversarial feature learning mechanism with the encoder network $E$. Given a set of $K$ seen types of adversarial examples $\tilde{X}=\{\tilde{X}_{1}, \tilde{X}_{2}, \cdots, \tilde{X}_{K}\}$, $D_A$ takes the encoded features $E(\tilde{X})$ as inputs and predicts attack types. The attack type of adversarial examples crafted by the $k$-$th$ attack is embodied as the attack-specific label $Y_k^p=\left[y_{k_1}^p,y_{k_2}^p, \ldots, y_{k_N}^p\right]^{\top}$, where $y_{k_n}^p=\left[\xi_1, \xi_2, \ldots, \xi_K\right]^{\top}$ is a one-hot vector and $\xi_i$ equals one when $i=k$ and zero otherwise. Based on the predictions about attack-specific labels, $D_A$ could reflect whether the encoded features contain attack-specific information. The objective function of $D_A$ is derived as follows: 
\begin{equation}
\label{eq1}
\mathcal{L}_{D_{A}}=-\frac{1}{K}\sum_{k=1}^{K} Y_{k}^{p} \cdot \log (\sigma(D_{A}(E(\tilde{X}_{k})))) \text{,}
\end{equation}
where $\sigma$ denotes the softmax layer.

In contrast, $E$ aims to remove the attack-specific information and make the learned encoded features indistinguishable for $D_A$. That is to say, as the adversary of $D_A$, $E$ is devoted to confuse $D_A$ from correctly predicting the attack-specific label and pushes its prediction close to the attack-confused label $Y_\zeta^p=\left[y_{\zeta_1}^p, y_{\zeta_2}^p, \ldots, y_{\zeta _N}^p\right]^{\top}$, where $y_{\zeta_n}^p=\left[1/K, 1/K, \ldots, 1/K\right]^{\top}$ is a $K$-dimensional constant vector. As a result, the objective of $E$ is as follows:
\begin{equation}
\label{eq2}
\mathcal{L}_{att}=-\frac{1}{K}\sum_{k=1}^{K} Y_{\zeta}^{p} \cdot \log (\sigma(D_{A}(E(\tilde{X}_{k})))) \text{.}
\end{equation}

\textbf{Normalization term:} 
Since the widespread types of attacks in the real world are very diverse, the adversarial examples used in the training procedure are often biased. Although the above adversarial feature learning mechanism could effectively defend against an unseen type of attack that is similar to seen types of attacks, this bias issue may lead a risk that the learned AIF has poor generalization ability for some unseen types of attacks whose perturbations are significantly different from those in seen types of attacks. For example, as shown in Figure~\ref{fig2}, adversarial 
noise crafted by pixel-constrained PGD and AA looks similar, while adversarial noise crafted by spatially-constrained STA presents significant difference from them. 

To address the bias issue, inspired by previous studies \cite{makhzani2015adversarial,larsen2016autoencoding,kingma2013auto}, we introduce a normalization term in the encoded space of AIF to decrease the undesirable risk. Specifically, the feature distribution of each type of attack $P_k(E(\tilde{X}_k))$ is matched to a multivariate Gaussian prior distribution $\mathcal{N}(0, I)$ through utilizing the \textit{Jensen-Shannon Divergence} (JSD). The normalization could make the encoded features of different types of adversarial examples have similar distributions, which is beneficial for robustly restoring natural examples in Section \ref{section3.3}. The JSD measure is the average of \textit{Kullback-Leibler divergences} between each distribution and the average distribution $\bar{P}$, which is formulated as $JSD\left(P_{1}, \cdots, P_{K}\right)=\frac{1}{K} \sum_{k=1}^{K} KL\left(P_{k} \| \bar{P}\right)$. In our method, the distribution of each encoded feature is expected to be similar to the uniform prior distribution. We replace $\bar{P}$ by $\mathcal{N}(0, I)$. The objective function of this normalization term is derived as:
\begin{equation}
\label{eq3}
\mathcal{L}_{nor}=JSD\left(P_{1}, \cdots, P_{K}\right)=\frac{1}{K} \sum_{k=1}^{K} KL\left(P_{k} \| \mathcal{N}\right) \text{.}
\end{equation}

\begin{figure*}[t]
\vskip 0.2in
\begin{center}
\centerline{\includegraphics[width= \textwidth]{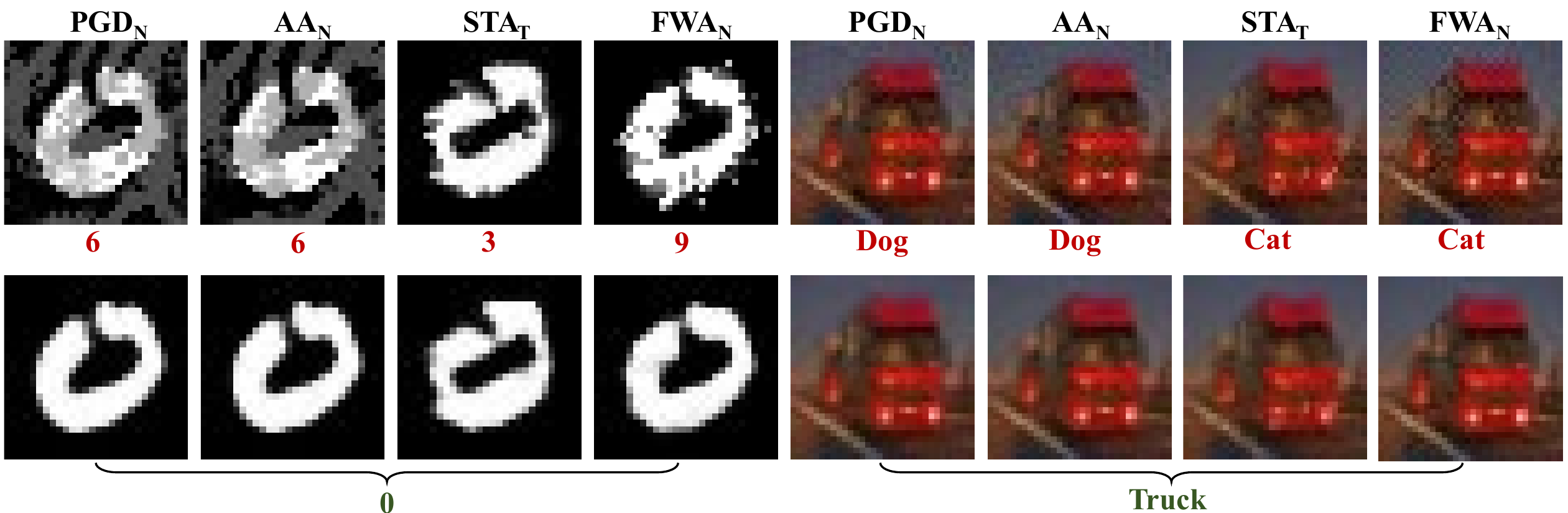}}
\caption{A visual illustration of the performance of our model against various attacks. (\textit{top:} adversarial examples; \textit{bottom:} restored examples). Subscripts ``N'' and ``T'' respectively indicate that the corresponding attacks are non-target and target attacks. PGD$_N$ is the seen type of attack while other attacks are regarded as unseen types of attacks.}
\label{fig4}
\end{center}
\vskip -0.2in
\end{figure*}

\subsection{Restoring Natural Examples}
\label{section3.3}
A hybrid objective function is also used to restore natural examples from AIF. The object consists of following two terms:

\textbf{Pixel similarity metric:} Adversarial noise could be viewed as the delicately crafted special noise. The widely used metric for image denoising or reconstruction would be able to achieve satisfactory results for generating examples close to natural examples \cite{shen2017ape}. Therefore, we apply the \textit{mean square error} (MSE) metric in the pixel space:
\begin{equation}
\label{eq4}
\mathcal{L}_{mse}=\sum_{k=1}^{K}\|G(E(\tilde{X}_{k}))-X\|_{2}^{2}\text{,}
\end{equation}
where $\|\cdot\|_{2}$ is the $L_2$ norm.

\textbf{Adversarial learning in pixel space:} As noted in \cite{2017Towards}, the decoder network based on MSE tends to synthesize blurry textures, which would lead to incorrect classification in the target model. To overcome the limitation, we introduce an image discriminator $D_I$ to form an adversarial mechanism in the pixel space with the decoder network $G$. $D_I$ is trained to identify natural examples $X$ as true data and identify synthesized examples as false data. We define the objective function of $D_I$ as:
\begin{equation}
\label{eq5}
\mathcal{L}_{D_{I}}=\sum_{k=1}^{K} [\log (D_{I}(G(E(\tilde{X}_{k}))))+\log (1-D_{I}(X))] \text{.}
\end{equation}
The adversarial objective function of $G$ is calculated as:
\begin{equation}
\label{eq6}
\mathcal{L}_{adv}=-\sum_{k=1}^{K} \log (D_{I}(G(E(\tilde{X}_{k})))) \text{.}
\end{equation}

\subsection{Implementation}
In order to make the encoded features invariant to different types of attacks and retain sufficient semantic classification information, we learn AIF by jointly optimizing the encoder network $E$ and the decoder network $G$. The overall objective function for $E$ is the combination of attack-invariant loss, normalization term loss and MSE loss: 
\begin{equation}
\label{eq7}
\mathcal{L}_{E}=\mathcal{L}_{mse}+\lambda_1 \mathcal{L}_{att} +\lambda_2 \mathcal{L}_{nor} \text{,}
\end{equation}
where $\lambda_1$ and $\lambda_2$ are positive parameters to trade off each component. The overall objective function for $G$ is given as: 
\begin{equation}
\label{eq8}
\mathcal{L}_{G}=\mathcal{L}_{mse}+\theta \mathcal{L}_{adv} \text{,}
\end{equation}
where $\theta$ is a trade-off parameter. Details of $\lambda_1$, $\lambda_2$, and $\theta$ are given in Section~\ref{section4.1}.

\begin{algorithm}[t]
   \caption{ARN: Adversarial Noise Removing Network}
   \label{alg1}
\begin{algorithmic}
   \STATE {\bfseries Input:} Natural examples $X$ and  adversarial examples $\tilde{X}$.
   \REPEAT
   \STATE 1: Sample a mini-batch $X_d$ and $\tilde{X}_d$ from X and $\tilde{X}$ res-\\ \quad pectively.
   \STATE 2: Forward-pass $\tilde{X}_d$ through $E$ to obtain encoded fea- \\ \quad  tures $E(\tilde{X}_d)$ and calculate $\mathcal{L}_{D_A}$ (Eq.~\ref{eq1}), $\mathcal{L}_{att}$ (Eq.~\ref{eq2}) \\ \quad and $\mathcal{L}_{nor}$ (Eq.~\ref{eq3}).
   \STATE 3: Forward-pass $E(\tilde{X}_d)$ through $G$ to restore natural \\ \quad examples and calculate $\mathcal{L}_{mse}$ (Eq.~\ref{eq4}), $ \mathcal{L}_{adv}$ (Eq.~\ref{eq6}) \\ \quad and $\mathcal{L}_{D_I}$ (Eq.~\ref{eq5}).
   \STATE 4: Back-pass and update $E$, $G$ to minimize $\mathcal{L}_{E}$ (Eq.~\ref{eq7}) \\ \quad and $\mathcal{L}_{G}$ (Eq.~\ref{eq8}).
   \STATE 5: Update $D_A$ and $D_I$ to minimize $\mathcal{L}_{D_A}$ (Eq.~\ref{eq1}) and \\ \quad $\mathcal{L}_{D_I}$ (Eq.~\ref{eq5}).
   \UNTIL {$E$ and $G$ converge.}
\end{algorithmic}
\end{algorithm}

The overall procedure is summarized in Algorithm 1. Given natural examples $X$ and adversarial examples $\tilde{X}$, we first sample a mini-batch $X_d$ and $\tilde{X}_d$ from X and $\tilde{X}$ respectively. 
Then, we forward-pass $\tilde{X}$ through $E$ to obtain encoded features $E(\tilde{X})$ and calculate $\mathcal{L}_{D_A}$ (Eq.~\ref{eq1}), $\mathcal{L}_{att}$ (Eq.~\ref{eq2}) and $\mathcal{L}_{nor}$ (Eq.~\ref{eq3}). Next, we forward-pass $E(\tilde{X})$ through $G$ and calculate $\mathcal{L}_{mse}$ (Eq.~\ref{eq4}), $ \mathcal{L}_{adv}$ (Eq.~\ref{eq5}) and $\mathcal{L}_{D_I}$ (Eq.~\ref{eq5}). Finally, we take a gradient step to update $E$, $G$, $D_A$ and $D_I$ to minimize $\mathcal{L}_{E}$ (Eq.~\ref{eq7}), $\mathcal{L}_{G}$ (Eq.~\ref{eq8}), $\mathcal{L}_{D_A}$ (Eq.~\ref{eq1}) and $\mathcal{L}_{D_I}$ (Eq.~\ref{eq5}). The above operations are repeated until $E$ and $G$ converge.

\begin{table*}[t]
\caption{Classification error rates (percentage) against adversarial examples crafted by pixel-constrained attacks on \textit{MNIST} and \textit{CIFAR-10} (\textit{lower is better}). `$\epsilon'$' means the raised  \textit{perturbation budget} $\epsilon$ of corresponding attack, it is set to 0.4 for \textit{MNIST} and 0.05 for \textit{\textit{CIFAR-10}}. `$7\times7$' denotes the size of sticker used by DOA.
For each attack we show the most successful defense with \textbf{bold} and the second result with \uline{underline}.}
\label{tab1}
\vskip 0.2in
\begin{center}
\begin{small}
\begin{sc}
\begin{tabular}{c|c|ccccccccc}
\toprule [1.0 pt]
\multirow{2}{*}{} &
  \multirow{2}{*}{\textbf{Defense}} &
  \multicolumn{9}{c}{\textbf{Attacks}} \\ \cline{3-11} 
 &       & None & PGD$_N$  & PGD$_T$       & CW$_N$   & DDN$_N$       & AA$_N$   & JSMA$_T$ & PGD$_{N\epsilon'}$ & AA$_{N\epsilon'}$ \\ \toprule [1.0 pt]
\multirow{10}{*}{LeNet} &
  None &
  \textbf{0.64} &
  100 &
  100 &
  100 &
  100 &
  100 &
  100 &
  100 &
  100 \\
 & $\text{AT}_{PGD}$    & 1.19      & 9.63  & 8.38       & 6.42  & 5.91       & 12.60  & 28.59 & 54.34   & 60.06  \\
 & DOA$_{7\times7}$   & 6.27      & 65.23 & 38.84      & 11.48 & 10.53      & 68.49 & 19.81 & 86.76   & 92.51  \\
 & $\text{APE-G}_{PP}$ & 1.57      & 8.76  & 3.20      & 2.34  & 2.15       & 12.40  & 36.49 & 34.86   & 46.72  \\
 & $\text{APE-G}_{DP}$ & 1.73      & 10.39  & 5.81       & 2.93  & 1.91       & 15.26  & 38.04 & 37.33   & 49.38  \\
 & $\text{HGD}_{PP}$ & 1.36      & \uline{1.89}  & \uline {1.30} & 1.67  & 1.54       & \uline{2.43}  & 50.62 & 75.79   & 90.34  \\
 & $\text{HGD}_{DP}$ & 1.18      & 2.56  & 1.91       & 1.79  & \uline {1.23} & 3.30  & 53.73 & 78.95   & 93.76  \\ \cline{2-11} 
 &
  $\text{ARN}_{PP}$ &
  1.16 &
  \textbf{1.85} &
  \textbf{1.29} &
  \textbf {1.45} &
  1.28 &
  \textbf{2.38} &
  \textbf{16.75} &
  \textbf{15.27} &
  \textbf{26.84} \\
 &
  $\text{ARN}_{DP}$ &
  \uline {1.11} &
  1.91 &
  1.80 &
  \uline{1.53} &
  \textbf{1.22} &
  2.97 &
  \uline {17.81} &
  \uline {17.63} &
  \uline {29.74} \\ \hline
\multirow{10}{*}{ResNet} &
  None &
  \textbf{7.67} &
  100 &
  100 &
  100 &
  99.99 &
  100 &
  100 &
  100 &
  100 \\
 & $\text{AT}_{PGD}$    & 12.86     & 51.02 & 49.68      & 50.17 & 49.19      & 53.66 & 44.59 & 59.09   & 61.65  \\
 & DOA$_{7\times7}$   & 9.82      & 89.03 & 73.96      & 24.11 & 49.29      & 97.52 & \textbf{23.26} & 96.83   & 97.75  \\
 & $\text{APE-G}_{PP}$ & 23.08     & 44.38 & 39.09      & 23.18  & 32.39      & 60.09 & 39.10 & 79.34   & 87.16  \\
 & $\text{APE-G}_{DP}$ & 24.23     & 45.96 & 41.50       & 27.43 & 24.73      & 64.82 & 41.67 & 83.19   & 89.92  \\
 & $\text{HGD}_{PP}$ & 10.41     & \uline{39.44} & 23.03      & 13.26 & 16.02      & 42.34 & 38.65 & 57.97   & 58.41  \\
 &
  $\text{HGD}_{DP}$ &
  9.42 &
  41.62 &
  25.30 &
  12.46 &
  \textbf{10.04} &
  43.45 &
  43.63 &
  58.63 &
  59.86 \\ \cline{2-11} 
 &
  $\text{ARN}_{PP}$ &
  8.21 &
  \textbf{38.66} &
  \textbf{20.43} &
  \textbf {11.47} &
  14.64 &
  \textbf{38.94} &
  \uline{35.49} &
  \textbf{49.45} &
  \textbf{52.64} \\
 &
  $\text{ARN}_{DP}$ &
  \uline {8.18} &
  40.28 &
  \uline {22.87} &
  \uline {12.24} &
  \uline {10.17} &
  \uline {41.27} &
  36.23 &
  \uline {52.87} &
  \uline {55.91}  \\ \toprule [1.0 pt]
\end{tabular}
\end{sc}
\end{small}
\end{center}
\vskip -0.2in
\end{table*}

\section{Experiments}
\label{section4}
In this section, we first introduce the datasets used in this paper (Section~\ref{section4.1}). We next show and analyze the experimental results of defending against pixel-constrained and spatially-constrained attacks on visual classification tasks, especially against adaptive attacks (Section~\ref{section4.2}). Finally, we conduct additional evaluations on the cross-model defense, ablation study and adversarial detection to further show the effectiveness of our ARN (Section~\ref{section4.3}).

\subsection{Experiment setup}
\label{section4.1}
\textbf{Datasets:} We verify the effective of our method on three popular benchmark datasets, i.e., \textit{MNIST} \cite{lecun1998gradient}, \textit{CIFAR-10} \cite{krizhevsky2009learning}, and \textit{LISA} \cite{jensen2016vision}. \textit{MNIST} and \textit{CIFAR-10} both have 10 classes of images, but the former contains 60,000 training images and 10,000 test images, and the latter contains 50,000 training images and 10,000 test images. 
To alleviate the problem of imbalance and extremely blurry data in \textit{LISA}, we picked 16 best quality signs with 3,509 training images and 1,148 test images from a subset which contains 47 different U.S. traffic signs \cite{eykholt2018robust,wu2019defending}. Adversarial examples are crafted by applying state-of-the-art attacks. These attacks can be divided into two categories: (i) \textit{Pixel-constrained attacks}, i.e., non-target $L_{\infty}$ norm PGD (PGD$_N$), target $L_{\infty}$ norm PGD (PGD$_T$), non-target DDN (DDN$_N$), non-target $L_{2}$ norm CW (CW$_N$), non-target AA (AA$_N$) and target JSMA (JSMA$_T$). (ii) \textit{Spatially-constrained attacks}, i.e., non-target STA (STA$_N$), target STA (STA$_T$), non-target FWA (FWA$_N$) and non-target RP$_2$ (RP$_N$). 

\textbf{Training details:} For fair comparison, all experiments are conduced on four NVIDIA RTX 2080 GPUs, and all methods are implemented by PyTorch. We use the implementation codes of PGD, DDN, CW, JSMA and STA in the \textit{advertorch toolbox} \citep{ding2019advertorch} and the implementation codes of RP2, FWA and AA provided by their authors. We set default perturbation budget $\epsilon=0.3$ and $\epsilon=8/255$ for \textit{MNIST} and \textit{CIFAR-10} respectively. More details about the attack approaches can be found in appendix A. Learning rates for the encoder network, the decoder network, the attack discriminator and the image discriminator are all set to $10^{-4}$, with the value of $\lambda_1$ = $10^{-1}$, $\lambda_2$ = $10^{-2}$, $\theta$ = $10^{-1}$ for \textit{MNIST}, the value of $\lambda_1$ = $10^{2}$, $\lambda_2$ = $10^{1}$, $\theta$ = $10^{2}$ for \textit{CIFAR-10} and \textit{LISA}. In addition, we consider the following deep neural network architectures as the target models: 
\begin{itemize}
\setlength{\itemsep}{0pt}
\setlength{\parsep}{0pt}
\setlength{\parskip}{0pt}
\item \textbf{MNIST:} The LetNet-5 architecture (LeNet) \cite{lecun1998gradient} embedded in the advertorch toolbox is used for the \textit{MNIST} digits recognition task.
\item \textbf{CIFAR-10:} The ResNet-110 (ResNet) architecture \citep{he2016deep}, the Wide-ResNet (WRN) architecture \cite{Zagoruyko2016WRN} and the VGG-19 (VGG) architecture \cite{simonyan2014very} are utilized for the classification task on \textit{CIFAR-10}. The depth and widen factors in WRN are set to $28 \times 20$.
\item \textbf{LISA:} We use the LISA-CNN architecture defined in \cite{eykholt2018robust} for the traffic sign recognition task according to the previous work \cite{wu2019defending}. The convolutional neural network contains three convolutional layers and one fully connected layer.
\end{itemize}

\subsection{Defense Results}
\label{section4.2}
\textbf{Defending against pixel-constrained attacks:} We select two attacks as seen types of attacks to craft adversarial examples, and use them together with natural examples as training data to train defense models. The other attacks are regarded as unseen types of attacks to evaluate the generalization ability of defense models. Considering that the $L_2$ norm distance between adversarial examples and natural examples varies greatly across different attacks, which may influence the performances of models, we construct two different combinations of seen types of attacks: (i) the target PGD and the non-target PGD (``$\text{defense}_{PP}$''). (ii) the non-target DDN and the non-target PGD (``$\text{defense}_{DP}$''). Figure~\ref{fig4} demonstrates that our ARN is effective to remove strong adversarial noise. Quantitative analysis in Table~\ref{tab1} represents that our ARN achieves better robust performance, especially reducing the success rate of JSMA$_T$ from 50.62\% to 16.75\% compared to previous state-of-the-art. Moreover, our ARN shows a significant improvement in defending against attacks with greater perturbation budgets (i.e., $\epsilon=0.4$ for \textit{MNIST} and $\epsilon=0.05$ for \textit{CIFAR-10}). 

\begin{table}[t]
\caption{Classification error rates (percentage) against adversarial examples crafted by spatially-constrained attacks (\textit{lower is better}). `$\epsilon'$' is set to 0.4 for \textit{MNIST} and 0.05 for \textit{CIFAR-10}. AP$_i$ denotes different adversarial patches crafted by RP$_2$ \cite{eykholt2018robust}. We show the most successful defense with \textbf{bold} and the second result with \uline{underline}.}
\label{tab2}
\begin{center}
\begin{small}
\begin{sc}
\begin{tabular}{p{0.03cm}|p{1.5cm}|p{0.55cm}p{0.55cm}p{0.55cm}p{0.55cm}p{0.9cm}}
\toprule [1.0 pt]
\multirow{2}{*}{}                                                              & \multirow{2}{*}{\textbf{Defense}} & \multicolumn{5}{c}{\textbf{Attacks}}          \\ \cline{3-7} 
 &       & None       & STA$_N$             & STA$_T$          & FWA$_N$      & FWA$_{N\epsilon'}$       \\ \toprule [1.0 pt]
\multirow{9}{*}{\begin{tabular}[c]{@{}c@{}}L\\ e\\ N\\ e\\ t\end{tabular}}     & None                              & \textbf{0.64} & 100 & 100 & 98.56  & 99.91 \\
 & AT$_{PGD}$    & 1.19       & 21.55          & 31.49          & 42.41        & 61.55          \\
 & DOA$_{7\times7}$   & 6.27       & 13.54          & 19.24          & \textbf{23.29}         & \textbf {40.16}    \\
 & APE-G$_{PP}$ & 1.57       & 16.57          & 21.40          & 33.95         & 51.81          \\
 & APE-G$_{DP}$ & 1.73        & 20.18          & 25.63          & 37.50         & 58.16           \\
 & HGD$_{PP}$ & 1.36       & 21.32          & 36.41          & 50.43         & 71.12          \\
 & HGD$_{DP}$ & 1.18       & 25.01          & 38.47          & 52.84         & 75.35          \\ \cline{2-7} 
 & ARN$_{PP}$ & 1.16       & \textbf{9.08}  & \textbf{13.73} & \uline {25.79}    & \uline{43.76}          \\
 & ARN$_{DP}$ & \uline {1.11} & \uline {10.14}     & \uline {14.51}    & 28.50  & 47.63 \\ \hline
\multirow{9}{*}{\begin{tabular}[c]{@{}c@{}}R\\ e\\ s\\ N\\ e\\ t\end{tabular}} & None                        & \textbf{7.67} & 100 & 100 & 99.83 & 99.98 \\
 & AT$_{PGD}$    & 12.86      & 44.86          & 44.60          & 40.32          & 49.37          \\
 & DOA$_{7\times7}$   & 9.82       & 38.33          & 28.02          & 49.69         & 62.00         \\
 & APE-G$_{PP}$ & 23.08      & 47.19          & 36.46             & 42.79         & 50.53          \\
 & APE-G$_{DP}$ & 24.23      & 49.93          & 37.51          & 45.26         & 57.61          \\
 & HGD$_{PP}$ & 10.41      & 42.89          & 31.97          & 37.67         & 43.41          \\
 & HGD$_{DP}$ & 9.42       & 49.52          & 36.06          & 35.95         & 42.87          \\ \cline{2-7} 
 & ARN$_{PP}$ & 8.21       & \textbf{36.81} & \textbf{23.62} & \textbf {24.17}   & \textbf {31.89}    \\
 & ARN$_{DP}$ & \uline {8.18} & \uline{37.74}    & \uline{26.90}     & \uline{27.10} & \uline{33.06} \\ \hline
\multirow{7}{*}{\begin{tabular}[c]{@{}c@{}}\\ C\\ N\\ N\\       \end{tabular}} &
  \textbf{} &
  None &
  AP$_1$ &
  AP$_2$ &
  AP$_3$ &
  AP$_4$ \\ \cline{2-7} 
 &
  None &
  \multicolumn{1}{l}{\textbf{0.86}} &
  \multicolumn{1}{l}{55.46} &
  \multicolumn{1}{l}{62.07} &
  \multicolumn{1}{l}{61.21} &
  \multicolumn{1}{l}{56.03} \\
 &
  AT$_{PGD}$ &
  \multicolumn{1}{l}{3.16} &
  \multicolumn{1}{l}{50.29} &
  \multicolumn{1}{l}{43.68} &
  \multicolumn{1}{l}{56.03} &
  \multicolumn{1}{l}{33.62} \\
 &
  APE-G &
  \multicolumn{1}{l}{3.43} &
  \multicolumn{1}{l}{\uline {8.33}} &
  \multicolumn{1}{l}{\uline {5.43}} &
  \multicolumn{1}{l}{21.56} &
  \multicolumn{1}{l}{24.71} \\
 &
  DOA$_{9\times5}$ &
  \multicolumn{1}{l}{2.59} &
  \multicolumn{1}{l}{18.39} &
  \multicolumn{1}{l}{6.90} &
  \multicolumn{1}{l}{25.86} &
  \multicolumn{1}{l}{\uline {8.91}} \\
 &
  DOA$_{7\times7}$ &
  \multicolumn{1}{l}{5.17} &
  \multicolumn{1}{l}{16.95} &
  \multicolumn{1}{l}{11.49} &
  \multicolumn{1}{l}{\uline {19.83}} &
  \multicolumn{1}{l}{10.06} \\ \cline{2-7} 
 &
  ARN &
  \multicolumn{1}{l}{\uline {2.31}} &
  \multicolumn{1}{l}{\textbf{5.46}} &
  \multicolumn{1}{l}{\textbf{3.74}} &
  \multicolumn{1}{l}{\textbf{6.90}} &
  \multicolumn{1}{l}{\textbf{6.03}} \\ \toprule [1.0 pt]
\end{tabular}
\end{sc}
\end{small}
\end{center}
\vskip -0.2in
\end{table}

\textbf{Defending against spatially-constrained attacks:} In addition to pixel-constrained attacks, some attacks focus on mimicking non-suspicious vandalism via spatial transformation and physical modifications \cite{gilmer2018motivating,wu2019defending}. We evaluate the robustness of above defense models against STA$_T$, STA$_N$ and FWA$_N$ on \textit{MNIST} and \textit{CIFAR-10}. As shown in Table~\ref{tab2}, our method achieves more effective defense and has better robustness. In particularly, our ARN significantly reduces the success rate of STA$_T$ from 36.41\% to 13.73\% in comparison to previous state-of-the-art, which has outstanding performance against pixel-constrained attacks. In order to further remove the spatially-constrained adversarial noise, we train our ARN by using adversarial examples crafted by PGD$_N$ and STA$_N$. The fooling rates of STA$_N$, STA$_T$ and FWA$_N$ on \textit{MNIST} are decreased from 9.08\%, 13.73\% and 25.79\% to 6.52\%, 7.94\% and 16.32\%, while the fooling rates of PGD$_N$ and PGD$_T$ are remained at 3.66\% and 2.51\% respectively. In addition, for protecting the target model on \textit{LISA}, defense models are trained based on two seen types of \textit{adversarial patches (AP)} crafted by RP$_2$, i.e., AP$_1$ and AP$_2$. Our ARN also achieves better performance on defending against unseen types of adversarial patches. The restored images are shown in appendix B.

\textbf{Leaked defenses:}
We study the following three different scenarios where defenses are leaked: 
\begin{itemize}
    \setlength{\itemsep}{0pt}
    \setlength{\parsep}{0pt}
    \setlength{\parskip}{0pt}
    \item[(i)] An attacker knows the per-processing defense model and directly uses white-box adaptive attacks \cite{carlini2017magnet} to break it. In this scenario, the attacker gains a copy of the trained defense model. The architecture and model parameters of the pre-processing model are both leaked to the attacker.
    \item[(ii)] An attacker trains a similar pre-processing defense model and then take the combination of the known pre-processing model and the original target model as a new target model to craft adversarial examples via gray-box adaptive attacks. We use APE-G$_{PP}$ and HGD$_{PP}$ as the known pre-processing models to craft adversarial examples via different types of attacks. 
    \item[(iii)] An attacker can utilize BPDA \citep{athalye2018obfuscated} strategy to bypass the pre-processing defense. Specifically, BPDA is different from the attack strategy which directly computes the gradient of the defense model $g(\cdot)$ and the target model $f(\cdot)$. If the knowledge of $g(\cdot)$ is inaccessible or if $g(\cdot)$ is neither smooth nor differentiable, $g(\cdot)$ cannot be backpropagated through to generate adversarial examples with a white-box attack that requires gradient signal. BPDA can approximate $\nabla_{x} f(g(x))$ by evaluating $\nabla_{x} f(x)$ at the point $g(x)$. This allows an attacker to compute gradients and therefore mount a white-box attack. BPDA is widely used to bypass pre-processing defenses. It can be used to explore whether an adversary can precisely approximate the gradient of the defense model for implementing white-box attacks. We combine BPDA with PGD$_N$ to evaluate our defense model. 
\end{itemize}

\begin{table}[t]
\caption{Classification error rates (percentage) against white-box and gray-box adaptive attacks (\textit{lower is better}) on \textit{MNIST}. ``TAR'' denotes the target attack and ``NON-TAR'' denotes the non-target attack. ``ITE-$\tau$'' means that the maximum number of iterations is controlled to be $\tau$ and ``$\epsilon$-$\tau$'' means that the perturbation budget is set to $\tau$. ``L'' denotes the original target model.}
\label{tab3}
\vskip 0.2in
\begin{center}
\begin{small}
\begin{sc}
\begin{tabular}{l|c|c|cc} \toprule [1.0 pt]
Target& \multirow{2}{*}{Attack} &
\multirow{2}{*}{Defense} & \multicolumn{2}{c}{Ite-40} \\ \cline{4-5} 
LeNet (L)  & &   & $\epsilon$-$0.3$    & $\epsilon$-$0.5$    \\ \toprule[1.0 pt]
APE-G+L & PGD$_T$       & APE-G     & 99.84 & 100 \\
HGD+L & PGD$_T$        & HGD  & 62.50 & 100 \\
ARN+L & PGD$_T$        & ARN     & 58.52 & 99.95 \\
APE-G+L & PGD$_T$        & ARN     & 1.49 & 3.80 \\ 
HGD+L & PGD$_T$        & ARN    & 2.56 & 10.65 \\ \toprule [1.0 pt]
Target & \multirow{2}{*}{Attack} & \multirow{2}{*}{Defense} & \multirow{2}{*}{Tar} & \multirow{2}{*}{Non-tar} \\
LeNet (L) & & & & \\ \toprule [1.0 pt]
APE-G+L & CW        & APE    &99.90 &98.07 \\ 
APE-G+L & CW        & ARN    &1.39 &1.28 \\ 
HGD+L & DDN        & HGD    &100 &100 \\ 
HGD+L & DDN        & ARN    &1.29 &1.42 \\ \toprule [1.0 pt] 
\end{tabular}
\end{sc}
\end{small}
\end{center}
\vskip -0.2in
\end{table}

\begin{table}[t]
\caption{Classification error rates (percentage) against BPDA (\textit{lower is better}). ``P+B'' denotes the hybrid attack of PGD$_N$ and BPDA \citep{athalye2018obfuscated}. ``L'' and ``R'' denote the original target models on \textit{MNIST} and \textit{CIFAR-10} respectively.}
\label{tab3_1}
\vskip 0.2in
\begin{center}
\begin{small}
\begin{sc}
\begin{tabular}{l|c|c|cc} \toprule [1.0 pt]
\multicolumn{5}{c}{MNIST: LeNet (L)} \\
Target & Attack & Defense & Ite-40 &Ite-100 \\ \toprule [1.0 pt]
APE-G+L & P+B & APE-G &72.01 &72.75 \\
ARN+L & P+B & ARN &24.65 &24.70 \\  \toprule [1.0 pt]
\multicolumn{5}{c}{CIFAR-10: ResNet-110 (R)} \\
Target & Attack & Defense & Ite-40 &Ite-100 \\ \toprule [1.0 pt]
APE-G+R & P+B & APE-G &89.06 &89.51 \\
ARN+R & P+B & ARN &60.47 &60.75 \\  \toprule [1.0 pt]
\end{tabular}
\end{sc}
\end{small}
\end{center}
\vskip -0.2in
\end{table}

As shown in Table~\ref{tab3} and Table~\ref{tab3_1}, experimental results present that our defense model achieves positive gains in these challenging settings compared to other pre-processing defenses. For example, the classification error rates against BPDA and white-box PGD$_T$ are decreased by 66\% and 24\% on average respectively. This may be due to the attack-invariant features being more robust against adversarial noise under the constraints of small perturbation budgets. The adversarial examples crafted by adaptive attacks and their restored examples are shown in appendix C. 

\begin{table}[t]
\caption{Classification error rates (percentage) of different target models with ARN (\textit{lower is better}) on \textit{CIFAR-10}. ARN is trained by using adversarial examples crafted against ResNet, and then is applied to WRN and VGG.}
\label{tab4}
\vskip 0.2in
\begin{center}
\begin{small}
\begin{sc}
\begin{tabular}{l|cllll}
\toprule [1.0 pt]
\multicolumn{1}{c|}{\multirow{3}{*}{Attack}} & \multicolumn{5}{c}{Target Model}                                                \\ \cline{2-6} 
\multicolumn{1}{c|}{} & \multicolumn{1}{c|}{ResNet} & \multicolumn{2}{c|}{WRN}                            & \multicolumn{2}{c}{VGG}                            \\ \cline{2-6}
\multicolumn{1}{c|}{} & \multicolumn{1}{c|}{ARN}    & \multicolumn{1}{c}{None} & \multicolumn{1}{c|}{ARN} & \multicolumn{1}{c}{None} & \multicolumn{1}{c}{ARN} \\ \toprule [1.0 pt]
PGD$_N$                                         & \multicolumn{1}{c|}{38.66} & 100   & \multicolumn{1}{l|}{33.38} & 100   & 36.15 \\
PGD$_T$                                         & \multicolumn{1}{c|}{20.43} & 99.91   & \multicolumn{1}{l|}{23.73} & 99.66   & 24.20 \\
CW$_N$                                          & \multicolumn{1}{c|}{11.47} & 100 & \multicolumn{1}{l|}{9.92} & 100 & 10.39 \\
DDN$_N$                                          & \multicolumn{1}{c|}{14.64} & 100 & \multicolumn{1}{l|}{9.20} & 100 & 9.57 \\
AA$_N$                                          & \multicolumn{1}{c|}{38.94} & 100   & \multicolumn{1}{l|}{36.94} & 100   & 37.59 \\
STA$_N$                                         & \multicolumn{1}{c|}{36.81} &100 & \multicolumn{1}{l|}{29.46} & 100 & 30.47 \\
STA$_T$                                         & \multicolumn{1}{c|}{23.62} & 99.95 & \multicolumn{1}{l|}{22.28} & 99.96 & 21.46 \\
FWA$_N$                                         & \multicolumn{1}{c|}{24.17} & 94.37 & \multicolumn{1}{l|}{24.23} & 95.21 & 23.31 \\ \toprule [1.0 pt]
\end{tabular}
\end{sc}
\end{small}
\end{center}
\vskip -0.2in
\end{table}

\subsection{Further Evaluations}
\label{section4.3}

\textbf{Cross-model defense results:}
In order to evaluate the cross-model defense capability of our ARN, we transfer the ARN$_{PP}$ model used for ResNet to other classification models, i.e., WRN and VGG. Results in Table~\ref{tab4} present that our ARN effectively removes adversarial noise crafted by various unseen types of attacks against WRN and VGG, which demonstrates that our ARN could provide generalizable cross-model protection.

\begin{figure}[t]
\vskip 0.2in
\begin{center}
\centerline{\includegraphics[width=2.8in]{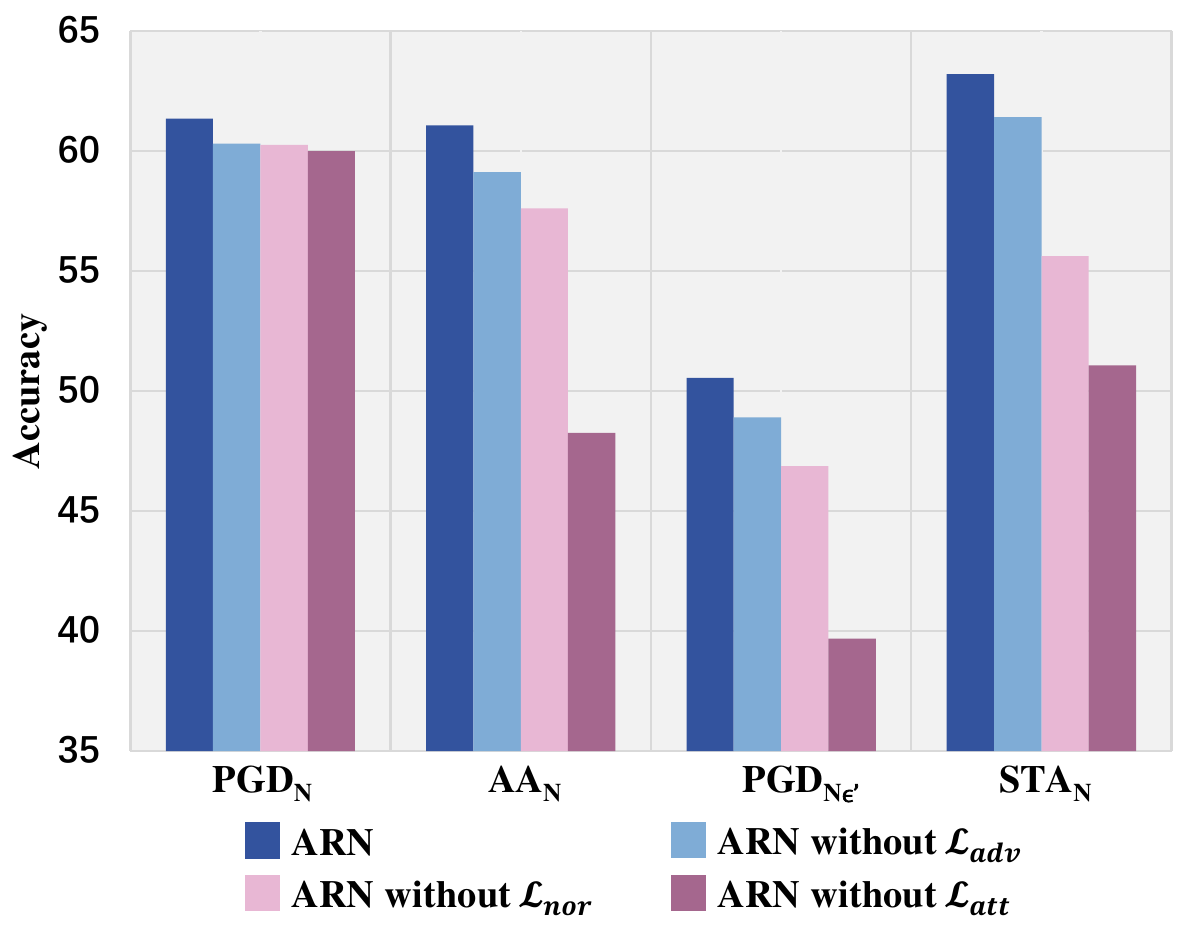}}
\caption{Ablation study. The figure shows the classification accuracy rates (percentage) of ResNet against different attacks (\textit{higher is better}) on \textit{CIFAR-10}. The performance of ARN against unseen STA$_N$ is significantly affected when $\mathcal{L}_{nor}$ is dropped. ARN trained without $\mathcal{L}_{att}$ has poor robust against unseen types of attacks i.e., AA$_N$, PGD$_{N\epsilon'}$ and STA$_N$. }
\label{fig5}
\end{center}
\vskip -0.3in
\end{figure}

\textbf{Ablation:} Figure~\ref{fig5} shows the ablation study on \textit{CIFAR-10}. We respectively remove the pixel adversarial loss $\mathcal{L}_{adv}$, the normalization term loss $\mathcal{L}_{nor}$ and the attack-invariant loss $\mathcal{L}_{att}$ to investigate their impacts on our ARN. We use PGD$_N$ and PGD$_T$ as seen types of attacks to train ARN. Removing $\mathcal{L}_{adv}$ slightly reduces the classification accuracy rates. The performance of ARN against STA$_N$ is significantly affected when $\mathcal{L}_{nor}$ is dropped. ARN trained without $\mathcal{L}_{att}$ no longer learns AIF and hence loses its superior generalizable ability to unseen types of attacks.  

\textbf{Adversarial examples detection:} Local intrinsic dimensionality (Lid) method \cite{ma2018characterizing} could distinguish between adversarial examples and natural examples by revealing the essential difference between them. In this way, we can evaluate our ARN from the perspective of detecting adversarial examples. A binary classifier is first trained to distinguish between positive examples (adversarial examples) and negative examples (natural examples). Then, we take the clean examples restored by our ARN as positive examples and input them to the classifier. The classifier presents low recall rates, i.e., 1.26\% for \textit{MNIST} and 8.48\% for \textit{CIFAR-10}, which reflects that restored examples are almost indistinguishable from natural examples. This demonstrates that our ARN could effectively remove adversarial noise. 

\textbf{Discussion on the number of seen types of attacks:}
In the above experiments, we choose two attacks as seen types of attacks to train the pre-processing model. Of course, the ideal number of seen types of attacks is not fixed. We think that the ideal number of seen types of attacks is related to the diversity of unseen types of attacks that may appear in a practical scenario. Specifically, If the number of unseen types of attacks is small or all unseen types of attacks are similar (e.g., CW$_N$ and CW$_T$ ), using one strong seen types of attacks (e.g., PGD$_N$) may be ideal. If the unseen types of attacks are quite different (i.e., CW$_N$, AA$_N$ and FWA$_N$), we can use more seen types of attacks (e.g., PGD$_N$ and STA$_N$) to train our defense model for providing robust protection. The selected seen types of attacks are expected to approximately cover the unseen types of attacks. In the real world, attacks are continuously evolving. The new attacks may have obvious discrepancies with previous attacks, and thus pose potential threats to the defense model. We could update the seen types of attacks and retrain the defense model to enhance the model’ s adversarial robustness.

\section{Conclusion}
\label{section5}
In this paper, we focus on designing a pre-processing model for adversarial defense against different unseen types of attacks. Inspired by cognitive science researches on the human brain, we propose an \textit{adversarial noise removing network} to restore natural examples by exploiting \textit{attack-invariant features}. Specifically, we introduce an adversarial feature learning mechanism to disentangle invariant features from adversarial noise. A normalization term is proposed in the encoded space of the invariant features to address the bias issue between the seen and unseen types of attacks. By minimizing the gap between the synthesized examples and natural examples, our method could restore natural examples from attack-invariant features. Experimental results demonstrate that our proposed model presents superior effectiveness against both pixel-constrained and spatially-constrained attacks, especially for unseen types of attacks and adaptive attacks. In future, we can extend the work in the following aspects. First, we can try to leak our defense model to attacks during the training process for improving the defense effective against adaptive attacks. Second, we can use the feedback of a target model (e.g. predictions of class labels) to train our defense model for further improving classification accuracy rates. Third, we can combine the pre-training model with recently proposed robust target model \cite{liu2015classification,xia2019anchor,xia2020part,wang2019improving,xia2021robust} to explore the robustness of the combined model against noisy data.


\section*{Acknowledgements}
DWZ was supported by the Fundamental Research Funds for the Central Universities and the Innovation Fund of Xidian University. TLL was supported by Australian Research Council Project DE-190101473. NNW, CLP and XBG were supported by the National Key Research and Development Program of China under Grant 2018AAA0103202 and the National Natural Science Foundation of China under Grants 62036007, 61922066, 61876142, 61772402, 61806152. BH was supported by the RGC Early Career Scheme No. 22200720, NSFC Young Scientists Fund No. 62006202 and HKBU CSD Departmental Incentive Grant. The authors thank the reviewers and the meta-reviewer for their helpful and constructive comments on this work.

\normalem
\bibliography{example_paper}
\bibliographystyle{icml2021}

\end{document}